\documentclass[runningheads]{llncs}

\usepackage[T1]{fontenc}
\usepackage[numbers]{natbib}
\usepackage{microtype}
\usepackage{nicefrac}
\usepackage{graphicx}
\usepackage{color}
\usepackage{booktabs} 
\usepackage{array}

\usepackage{hyperref}

\usepackage{amsmath}
\usepackage{mathtools}
\usepackage{amssymb}

\usepackage[capitalize,noabbrev]{cleveref}

\usepackage{macros}
\usepackage{nicefrac}
\usepackage{dsfont}
\usepackage{caption}
\usepackage{subcaption}

\begin{document}
\title{Learning Action Embeddings \\for Off-Policy Evaluation}
\titlerunning{Learning Action Embeddings for Off-Policy Evaluation}
%

\author{Matej Cief\inst{1, 2}\orcidID{0000-0001-9225-5155} \and
Jacek Golebiowski\inst{3}\orcidID{0000-0001-8053-8318} \and
Philipp Schmidt\inst{3}\and
Ziawasch Abedjan\inst{4}\orcidID{0000-0002-2846-1373} \and
Artur Bekasov\inst{5}}
%
\authorrunning{Cief et al.}
%
\institute{Brno University of Technology, Brno, Czech Republic\footnote{Work done during an internship at Amazon.}\and
Kempelen Institute of Intelligent Technologies, Bratislava, Slovakia
\email{matej.cief@kinit.sk}\and
Amazon, Berlin, Germany\and
Leibniz University Hannover, Hanover, Germany\and
Amazon, London, United Kingdom}

\maketitle              
\begin{abstract}
    Off-policy evaluation (OPE) methods allow us to compute the expected reward of a policy by using the logged data collected by a different policy. 
    However, when the number of actions is large, or certain actions are under-explored by the logging policy, existing estimators based on inverse-propensity scoring (IPS) can have a high or even infinite variance.
    \citet{saito_off-policy_2022} propose marginalized IPS (MIPS) that uses action \emph{embeddings} instead, which reduces the variance of IPS in large action spaces.
    MIPS assumes that good action embeddings can be defined by the practitioner, which is difficult to do in many real-world applications.
    In this work, we explore \emph{learning} action embeddings from logged data.
    In particular, we use intermediate outputs of a trained reward model to define action embeddings for MIPS.
    This approach extends MIPS to more applications, and in our experiments improves upon MIPS with pre-defined embeddings, as well as standard 
    baselines, both on synthetic and real-world data.
    Our method does not make assumptions about the reward model class, and supports using additional action information to further improve the estimates.
    The proposed approach presents an appealing alternative to DR for combining the low variance of DM with the low bias of IPS.

\end{abstract}
\section{Introduction}
The \emph{multi-armed bandit} (MAB) framework is commonly used to model the interaction between recommender systems or search engines with their users ~\citep{zhou_survey_2016}. In MAB, an \emph{agent} performs \emph{actions} and receives \emph{rewards}, where each reward is sampled from an unknown reward distribution conditioned on the action. 
The goal is to learn a \emph{policy} for the agent that maximizes the cumulative reward over multiple iterations.
In a \emph{contextual} MAB, the reward distribution is also conditioned on a \emph{context} that is observed by the agent, and is used as an additional input to the policy function.

Bandit policies are commonly learned \emph{online}, where we run a policy in production and update it iteratively using the observed rewards~\citep{zhou_survey_2016}. 
However, deploying untested policies to production is risky.
An alternative is to leverage the logged data from historical customer interactions to learn a policy \emph{offline}.
The fundamental problem of offline learning is \emph{off-policy evaluation} (OPE), where we try to estimate the expected reward of a new policy without deploying it.
The disadvantage of OPE is that the computed value is only an \emph{estimate} of the true value of the policy, and the success of offline learning depends on how accurate this estimate is. 
When the policy used to gather logged data has a low probability of choosing a subset of actions, existing estimators based on \emph{inverse propensity scoring}~\citep[IPS,][]{robins_estimation_1994} can be inaccurate \citep{sachdeva_off-policy_2020}.
This problem is especially potent when the number of actions is large, as is often the case in recommender systems~\citep{swaminathan_off-policy_2017}.

Embedding contexts into a lower-dimensional space can improve the off-policy estimates when dealing with a large number of contexts~\citep{zhou_survey_2016}. 
Similarly, \citet{saito_off-policy_2022} propose to embed the actions, and to use the embeddings in \emph{marginalized} IPS (MIPS).
\citeauthor{saito_off-policy_2022} use additional action information to define the embeddings.
For example, in fashion recommendation, where an action is a particular fashion item to recommend, they use the price, the brand and the category of an item as embedding dimensions.
For many problems of interest defining action embeddings by hand will be difficult, and/or will require expert knowledge. 

Motivated by the two considerations above, we propose methods for \emph{learning} or \emph{fine-tuning} the action embeddings, aiming to improve the performance of MIPS. 
Our key hypothesis is that an action embedding that is useful for reward prediction will also be useful for MIPS.
In particular, we propose to use the intermediate outputs of a trained reward model as action embeddings for MIPS, instead of using them for direct reward prediction as in DM.

We study this approach on both synthetic and real-world datasets, demonstrating that it consistently outperforms MIPS with pre-defined action embeddings, as well as common baselines like standard IPS, \emph{direct method} (DM), and \emph{doubly robust} estimation \citep[DR,][]{dudik_doubly_2014}.
We demonstrate that the method is not sensitive to reward model misspecification: a linear regression reward model produces useful action embeddings even when the true reward function itself is non-linear.
We propose methods for utilizing additional action information when it is available, and show that learning/fine-tuning action embeddings is especially useful when the additional action information is high-dimensional, or when the information affecting the reward is incomplete.
Finally, we provide a theoretical analysis of our method, showing that it can be interpreted as kernelized DM.

\section{Background and related work}
\label{sec: background}

The interaction of a recommender system with the user starts with the system receiving the context information (e.g.\ user's purchase history) $x \in \cX \subseteq \mathbb{R}^{d_\cX}$ drawn i.i.d.\ from an unknown distribution $p(x)$. 
The system then chooses an action (e.g.\ a ranked set of products) $a \sim \pi(a \mid x)$ from the set of available actions $\cA$ according to the system's policy $\pi$.
Finally, the user interacts with the displayed products via clicks or purchases, and the system receives a reward $r \in [r_{min}, r_{max}]$ sampled from a reward distribution $p(r \mid x, a)$. 
This process repeats for $n$ rounds, where in round $t$ the system observes a context $x_t$, draws an action $a_t$ according to $\pi(a \mid x_t)$, and observes a reward $r_t \sim p(r \mid x_t,a_t)$.
All interactions are recorded in a \emph{logged dataset} $\cD = \set{(x_t, a_t, r_t)}_{t=1}^n$. 
The policy $\pi_0$ that collects $\cD$ is called a \emph{logging policy}.
The goal of \emph{off-policy evaluation} \citep[OPE,][]{dudik_doubly_2014, wang_optimal_2017, su_cab_2019, su_doubly_2020, metelli_subgaussian_2021, saito_off-policy_2022} is to develop $\hat{V}$, an estimator of the policy \emph{value} $V(\pi) = \mathbb{E}_{p(x)\pi(a \mid x)p(r\mid x,a)}[r]$ such that\ $\hat{V}(\pi)~\approx~V(\pi)$.

\emph{Inverse propensity scoring} \citep[IPS,][]{robins_estimation_1994} is a commonly used estimator that works by re-weighting the rewards in $\cD$ based on how likely the corresponding actions are to be selected under the target policy:
\begin{align}
\hat{V}_{\text{IPS}}(\pi) = \frac{1}{n}\sum_{t=1}^n\frac{\pi(a_t \mid x_t)}{\pi_0(a_t \mid x_t)}r_t.
\end{align}
Many state-of-the-art OPE methods are based on IPS \citep{dudik_doubly_2014,wang_optimal_2017,su_cab_2019,su_doubly_2020,metelli_subgaussian_2021}. 
A key assumption of IPS, however, is that the logging policy $\pi_0$ and the target policy $\pi$ have \emph{common support}, i.e.\ $\pi(a \mid x) > 0 \implies \pi_0(a \mid x) > 0$ \citep{saito_off-policy_2022}. 
When dealing with large action spaces, it can be difficult to select a logging policy that puts non-zero probability on \emph{all} actions, yet does not degrade the customer experience.
Even if the two policies have common support, a low probability of selecting an action under the logging policy results in a large weight $\nicefrac{\pi(a_t \mid x_t)}{\pi_0(a_t \mid x_t)}$, which increases the variance of the IPS estimate.
To reduce the variance of IPS, we can either clip the importance weights according to a tunable clipping parameter \citep{swaminathan_counterfactual_2017} or self-normalize them \citep[SNIPS,][]{swaminathan_self-normalized_2015}, but the former biases the estimator and the latter can still result in a relatively high variance.
To deal with continuous space, \citet{kallus_policy_2018} use a kernel function to calculate propensities for IPS. 
We also use a similar technique to estimate the propensities of learned embeddings, and we show that the propensity estimation used in MIPS can be interpreted as kernel regression in \cref{sec: analysis}.

In their work \citet{saito_off-policy_2022} instead assume that useful \emph{action embeddings} are available to the estimator, and that we can pool information across similar actions to improve the estimates.
Suppose we have an action embedding $e \in \cE \subseteq \mathbb{R}^{d_\cE}$ for each action $a$. 
\citeauthor{saito_off-policy_2022} show that if the logging policy $\pi_0$ has common \emph{embedding support}, i.e.\ $p(e \mid \pi, x) > 0 \implies p(e \mid \pi_0, x) > 0$, and the action $a$ has no \emph{direct effect} on the reward $r$, i.e.\ $a \perp r \mid x,e$, then the proposed \emph{marginalized} IPS (MIPS) estimator
\begin{align}
    \label{eq: mips estimator}
    \hat{V}_{\text{MIPS}}(\pi) = \frac{1}{n}\sum_{t=1}^n\frac{p(e_t \mid \pi, x_t)}{p(e_t \mid \pi_0, x_t)}r_t
\end{align}
is unbiased.
Experiments by \citeauthor{saito_off-policy_2022} demonstrate that in practice MIPS has a lower variance than IPS, and produces estimates with a lower MSE.
Our goal in this work is to develop a method for \emph{learning} the action embeddings from logged data to further improve MIPS, and to make it applicable to problems where pre-defined embeddings are not available.

The \emph{direct method} (DM) learns the model of the expected reward for the context-action pair, and uses it to estimate the expected reward of a policy.
DM has a low variance, but suffers from a high bias when the model is misspecified.
\emph{Doubly robust} estimator \citep[DR,][]{dudik_doubly_2014,farajtabar_more_2018,wang_optimal_2017} combines DM and IPS: it uses the learned model of the reward as a control variate.
In other words, it computes a non-parametric estimation on the reward model residuals.
If the expected reward is correctly specified, DR can achieve a lower variance than IPS, and a lower bias than DM.
Several extensions of DR have been proposed \citep{wang_optimal_2017, su_doubly_2020}, but, to the best of our knowledge, none of them aim to improve the performance of DR in large action spaces.

Two methods for off-policy evaluation with large action spaces have been developed in parallel with our work \citep{peng_offline_2023, saito_off-policy_2023}.
\citet{peng_offline_2023} propose to cluster similar actions, and replace individual action propensity scores with those of action clusters.
\citet{saito_off-policy_2023} train a two-step regression model, hypothesizing that similar actions share a cluster effect on the reward, but also have their own residual effect. 

\section{Methods}
\label{sec: methods}
In this work we propose to use the reward signal to learn the embeddings of actions for the MIPS estimator, which would estimate the propensities of these embeddings, and use them instead of the original action propensities to re-weight the observed rewards.

\subsection{Motivation}

We first formalize the main assumption that motivates the proposed method. 
\begin{assumption}[Action Embedding Space]
\label{assumption: action embedding space}
    For every set of actions $\cA$, there exists a lower-dimensional embedding space $\cE \subseteq \mathbb{R}^{d_\cE}$ so that every action $a \in \cA$ can be mapped to an embedding $e \in \cE$ while $p(r \mid a, x) \approx p(r \mid e, x)$ holds for every context $x$. 
    We denote $d_\cE < \abs{\cA}$ to be the number of dimensions of embedding space $\cE$.
\end{assumption}
This assumption holds when $d_\cE = \abs{\cA}$ as the actions can be mapped as one-hot encoded representations. 
But we hypothesize that in practice some actions are ``similar'', and that the action space can be represented more efficiently. 
For example, if certain types of customers mostly clicked on shoes in the logged data, we expect to observe a high reward for a shoe product, even if we have never recommended this particular product for this type of customer before.

Based on Assumption \ref{assumption: action embedding space}, the environment can be represented as a graphical model $a \to e \to r \gets x$.
This can be learned the same way as learning a reward model for DM, but instead of using the model for reward prediction, we use the intermediate model outputs as action embeddings. 
As model misspecification is one of the biggest issues with DM \citep{farajtabar_more_2018}, it is often difficult to predict the reward end-to-end. 
But even when the true reward function is complex, and the model class is not rich enough to recover it, we hypothesize that the model will learn a useful $a \to e$ mapping that can be used in MIPS as defined in \cref{eq: mips estimator}. 

To test our hypothesis, we design the following toy example\footnote{Code to reproduce this and further experiments is available at \url{https://github.com/amazon-science/ope-learn-action-embeddings}.}.
Let $\cD = \set{(x_t, a_t, r_t)}_{t=1}^n$ for $n = 1000$ be the logged dataset, where a context $x \in [-1, 1]^{d_\cX},\, d_\cX = 5$, is drawn from the standard normal distribution, action $a \sim \pi_0$ is chosen from set $\cA$ by the unconditional logging policy $\pi_0(a) = \nicefrac{\nu_a}{\sum_{a' \in \cA} \nu_{a'}}$, where $\nu_a \sim \text{Exp}(1)$ is drawn from an exponential distribution, and the reward is generated by $r = f_a(x) + \eta$, where $f_a(x) = 1 / (1 + e^{-(x\T\theta_a+\mu_a)})$ is a logistic model for action $a$ with randomly initialized parameters $\theta_a \sim \cN(0, 1)^{d_\cX}, \mu_a \sim \cN(0, 1)$, and $\eta \sim \cN(0, 0.1)$ is random noise.
We use the uniform random target policy.
For DM, we fit a linear regression $\hat{f}_a(x)$ to estimate $r$. 
Of course, linear regression is not expressive enough to fit the non-linear $f_a$, but it learns the correct \emph{order} of the contexts, meaning $\forall x_1, x_2 \in \cX, \hat{f}_a(x_1) < \hat{f}_a(x_2) \Leftrightarrow f_a(x_1) < f_a(x_2)$.
We take these learned embeddings and use them in the MIPS estimator \citep{saito_open_2021} and refer to this method as \emph{Learned MIPS}. 
We vary the number of actions $\abs{\cA} \in \set{50, 100, 200, 500, 1000}$, while keeping the size of the logged data fixed. 
For each $\abs{\cA}$ we synthesize 50 reward functions, and for each reward function we generate 15 logged datasets to evaluate the methods. 
The results are summarized in \cref{table: example}.

\begin{table}[t]
\caption{Mean squared error of the estimators on a synthetic experiment while varying the number of actions. Reporting mean and standard error averaged over 50 runs.}
\vskip -0.1in
\label{table: example}
\begin{center}
\begin{small}
\begin{sc}
\begin{tabular}{L{0.05\textwidth}C{0.2\textwidth}C{0.2\textwidth}C{0.2\textwidth}}
\toprule
$\abs{\cA}$ & IPS & DM & Learned MIPS \\
\midrule
50   & $0.63\pm0.05$ & $0.64\pm0.05$ & $\mathbf{0.56\pm0.05}$\\
100  & $0.68\pm0.04$ & $2.13\pm0.11$ & $\mathbf{0.58\pm0.05}$\\
200  & $0.61\pm0.04$ & $7.16\pm0.19$ & $\mathbf{0.55\pm0.04}$\\
500  & $0.71\pm0.05$ & $27.9\pm0.32$ & $\mathbf{0.56\pm0.05}$\\
1000 & $0.70\pm0.04$ & $62.3\pm0.31$ & $\mathbf{0.55\pm0.03}$\\
\bottomrule
\end{tabular}
\end{sc}
\end{small}
\end{center}
\vskip -0.25in
\end{table}

As the number of samples per action decreases, the learned embeddings become increasingly inaccurate for \emph{direct} reward prediction (see the increasing error of DM in \cref{table: example}), but they still reflect the structure of the problem, and we can build an accurate model-free estimator on top of them (Learned MIPS outperforms both DM and IPS).
This simple example demonstrates the potential of the proposed method as a novel way to combine the strengths of model-based and model-free off-policy estimation methods.


\subsection{Algorithm}

\begin{figure}[t]
    \centering
    \includegraphics{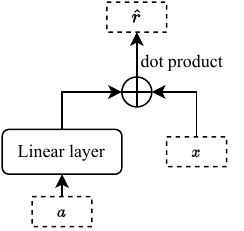}
    \caption{The model we use to learn the action embeddings. 
    We use a simple linear model in this work, but any model class can be used. 
    Input $a$ can be a one-hot encoded representation of the action identity, a pre-defined action embedding, or a concatenated vector of both.}
    \label{fig: model}
    \vskip -0.15in
\end{figure}

As shown in the previous section, we can learn action embeddings by fitting a separate linear regression model $\hat{f}_a(x)$ for each action $a$.
In general, as in DM, we can use any model class.
For example, we can fit a deep neural network, and use the intermediate outputs as action embeddings.
In this work, we use a linear model visualized in \cref{fig: model}.
In our preliminary experiments, more complex neural network models demonstrated comparable or worse performance.
We hypothesize that this is due to the high expressivity (and hence high variance) of neural networks, which is counter to our goal of reducing estimator variance. 
We leave further empirical study and theoretical analysis of this phenomenon for future work.


The model takes the action identity $a$, and computes a reward estimate $\hat{r}$ as a dot product between the action embedding and the context vector $x$. 
We train the model to minimize the MSE between $\hat{r}$ and the reward $r$ observed in the logged data.
The learned embedding for action $a$ is the output of the linear layer.
As we show in \cref{sec: evaluation}, we can also include additional action information (e.g.\ content-based features of the corresponding product), which can further improve the performance of the estimator.

Having learned the action embeddings, we follow MIPS \citep{saito_off-policy_2022} and fit a logistic regression model to estimate $p(a \mid e)$.
This estimate is used to compute the embedding propensities ($p(e \mid \pi, x)$ and $p(e \mid \pi_0, x)$) in \cref{eq: mips estimator}, noting that $p(e \mid \pi, x) = p(a \mid e)\pi(a \mid x)$.
We discuss this step in more detail in the next section, where we show a connection between our method and kernelized DM.

\subsection{Connection to DM}
\label{sec: analysis}

Learned MIPS resembles DM in that both methods solve the OPE task by first training a reward predictor.
The two methods differ, however, in how they apply said predictor.
In this section, we show a direct connection between learned MIPS and DM.
In particular, we show that we can interpret the proposed method as DM that uses \emph{kernel regression} with a learned feature embedding as the reward predictor.

To simplify the notation, we consider a non-contextual estimation task: given logged data $\cD = \set{a_t, r_t}_{t=1}^n$ gathered with a policy $\pi_0(a)$, we want to estimate the value of a policy $\pi(a)$. 
In this setting, the DM estimate is simply
\begin{align}
    \hat{V}_\text{DM}(\pi) = \sum_{a \in \cA} \pi(a)\hat{r}(a),
\end{align}
where $\hat{r}(a)$ is the model of expected reward for action $a$ that we learn using $\cD$ as the training data.
We now define $\tilde{\cD} = \set{a_t, e_t, r_t}_{t=1}^n$, where $e_t \sim p\left(e \mid a_t\right)$ is the corresponding action embedding.
The MIPS estimate is then
\begin{align}
    \hat{V}_\text{MIPS}(\pi) 
    &= \frac{1}{n}\sum_{t=1}^n\mathbb{E}_{p(a \mid e_t)} \left[\frac{\pi(a)}{\pi_0(a)}\right]r_t
    = \frac{1}{n}\sum_{t=1}^n\sum_{a \in \cA}\frac{\pi(a)}{\pi_0(a)}p(a \mid e_t)r_t\\
    &= \sum_{a \in \cA}\pi(a)\frac{1}{n}\sum_{t=1}^n\frac{p(a \mid e_t)}{\pi_0(a)}r_t
    = \sum_{a \in \cA}\pi(a) \tilde{r}(a), \label{eq:mips_as_dm}
\end{align}
where $p(a \mid e_t)$ is estimated as in \cref{eq:lda_pred}.
In other words, the MIPS estimate matches the DM estimate with a different expected reward model:
\begin{align}
    \tilde{r}(a) = \frac{1}{n}\sum_{t=1}^n\frac{p(a \mid e_t)}{\pi_0(a)}r_t. \label{eq:mips_r_estimate}
\end{align}
We now try to understand the nature of $\tilde{r}$.
Intuitively, ${p(a \mid e)}$ will be higher when the embedding of $a$ is ``closer'' to $e$ according to some distance measure. 
In particular, let $e \sim \cN(f(a), \sigma^2I)$ be the embedding distribution, where $f(a)$ is the learned deterministic embedding.
As in MIPS, we estimate ${p(a \mid e)}$ by fitting a probabilistic classifier to the dataset $\set{a_t, e_t}_{t=1}^n$.
\citeauthor{saito_off-policy_2022} use logistic regression, but here we consider \emph{linear discriminant analysis} \citep[LDA,][Section 4.3]{hastie_elements_2009}, a comparable model.
LDA fits a normal distribution to each class,
hence we expect it to recover $\cN(f(a), \sigma^2I)$.
We also expect it to recover $\pi_0(a)$ as the class prior, because we sampled $a_t \sim \pi_0(a)$ to produce the training data $\set{a_t, e_t}_{t=1}^n$.
The predictive probability of LDA is then
\begin{align}
    p(a\mid e) \propto \cN(e;f(a),\sigma^2I)\pi_0(a). \label{eq:lda_pred}
\end{align}

Combining \cref{eq:mips_r_estimate,eq:lda_pred} we get
\begin{align}
    \tilde{r}(a) \propto \frac{1}{n}\sum_{t=1}^n\cN(e_t;f(a),\sigma^2I)r_t, \label{eq:lda_exp_reward}
\end{align}
where
\begin{align}
    \cN(e_t;f(a),\sigma^2I) \propto \exp\left(-\frac{1}{2\sigma^2}(e_t-f(a))\T(e_t-f(a))\right).
\end{align}
This confirms our intuition: if we use LDA as the classifier, the expected reward $\tilde{r}(a)$ in \cref{eq:mips_as_dm} is a weighted combination of all rewards in the logged data, where samples with $e_t$ ``closer'' to the action embedding $f(a)$ have a higher weight.
For the normal embedding distribution, the weight is determined by a squared Euclidean distance between vectors $e_t$ and $f(a)$, but other distributions will induce different distance functions.
In fact, \cref{eq:lda_exp_reward} is equivalent to \emph{kernel regression} \citep[Chapter~10]{dhrymes_topics_1989} in a learned embedding space determined by $f$, with a kernel determined by the embedding distribution.


While the above holds when using LDA as the classifier, in our experiments we follow \citeauthor{saito_off-policy_2022} and use the logistic regression classifier.
Both of these methods fit linear decision boundaries and give similar results in practice
(\citealp{efron_efficiency_1975}; \citealp[Chapter~4]{hastie_elements_2009}), which means that our interpretation likely holds for logistic regression-based MIPS as well. 
As DM, our method does not provide theoretical guarantees on the bias/variance for a general reward model class, but, as we show in the next section, demonstrates good performance in practice.
\section{Empirical evaluation}
\label{sec: evaluation}

In this section we extend our proof-of-concept experiments in \cref{sec: methods} to a more complex synthetic dataset, and conduct experiments on a real-world fashion recommendation dataset. 
In the synthetic experiments, we evaluate two classes of methods based on whether the method uses additional action information or not.
We find that our methods outperform all the standard baselines and, in some cases, improve upon the true embeddings used to generate the reward.

\subsection{Synthetic experiments}
\label{sec: synthetic experiments}

For our empirical evaluation, we follow the setup by \citet{saito_off-policy_2022}. 
We generate synthetic data by first sampling $d_\cX$-dimensional context vectors $x$ from the standard normal distribution. 
We also sample $d_\cE$-dimensional categorical action embeddings $e \in \cE$ from the following distribution
\begin{align}
    \label{eq: embedding from action}
    \displaystyle p(e \mid a) = \prod_{k=1}^{d_\cE}\frac{\exp{\alpha_{a, e_k}}}{\sum_{e'\in\cE_k}\exp{\alpha_{a, e'}}},
\end{align}
where $\set{\alpha_{a, e_k}}$ is a set of parameters sampled independently from the standard normal distribution. 
Each dimension of $\cE$ has a cardinality of 10. 
We then synthesize the expected reward as 
\begin{align}
    \label{eq: expected reward}
    q(x, e) = \sum_{k=1}^{d_\cE}\eta \cdot (x\T Mx_{e_k}+\theta\T_xx+\theta\T_ex_{e_k}),
\end{align}
where $M$ is the parameter matrix and $\theta_x$, $\theta_e$ are parameter vectors sampled uniformly in $[-1, 1]$. 
$x_{e_k}$ is a parameter vector corresponding to the $k$-th dimension of the action embedding and is sampled from the standard normal distribution. 
Parameter $\eta_k$ specifies the importance of the $k$-th dimension and is drawn from a Dirichlet distribution $\eta \sim \text{Dir}(\alpha)$ with its parameters $\alpha = (\alpha_1,\dots,\alpha_{d_\cE})$ sampled from a uniform distribution $[0, 1)$. 
With this setup, the action embeddings are fully informative, and the action identity has no \emph{direct effect} on the reward. 
On the other hand, if action embeddings have some missing dimensions, they do not fully capture the action's effect on the reward.
In this case, we hypothesize that we can learn better embeddings, and we test this in one of the experiments.

We also follow \citet{saito_off-policy_2022} in how we synthesize the logging policy $\pi_0$. 
We apply the softmax function to $q(x, a) = \mathbb{E}_{p(e\mid a)}[q(x, e)]$ as 
\begin{align}
    \label{eq: logging policy}
    \pi_0(a \mid x) = \frac{\exp{\beta \cdot q(x,a)}}{\sum_{a'\in\cA}\exp{\beta \cdot q(x,a')}},
\end{align}
where $\beta$ is the parameter that controls the entropy and the optimality of the logging policy. 
A large positive value leads to an optimal, near-deterministic logging policy, while 0 leads to a uniform-random policy. 
We define the target policy $\pi$ as an $\eps$-greedy policy
\begin{align}
    \pi(a \mid x) = (1 - \eps) \cdot \I{a=\argmax_{a'\in\cA}a(x,a')} + \frac{\eps}{\abs{\cA}},
\end{align}
where $\eps \in [0, 1]$ controls the exploration rate of the target policy. 
Similar to \citet{saito_off-policy_2022}, we set $\beta = -1$ and $\eps = 0.05$ in our experiments.

To generate a batch of logged data, we follow an iterative process to generate $n$ samples  $\cD = \set{x_t, a_t, e_t, r_t}_{t=1}^n$. 
In each iteration, we sample a context $x$ from the standard normal distribution ($d_\cX = 10$), and a discrete action $a$ from $\pi_0$ according to \cref{eq: logging policy}.
Given an action $a$, we then sample a categorical action embedding $e$ using~\cref{eq: embedding from action}. 
Finally, we sample the reward from a normal distribution with mean $q(x, e)$ defined in \cref{eq: expected reward} and standard deviation $\sigma = 2.5$~\citep{saito_off-policy_2022}.

We assume three variants of our method, based on which action representation is passed to the model:
\begin{itemize}
    \itemsep=0em
    \item \emph{Learned MIPS OneHot} only sees one-hot encoded action identities.
    \item \emph{Learned MIPS FineTune} only sees pre-defined action embeddings.
    \item \emph{Learned MIPS Combined} sees both of the above concatenated into a single vector.
\end{itemize}

We compare our methods to all common estimators, including DM, IPS, and DR. 
To gauge the quality of the learned embeddings, we specifically compare to MIPS with pre-defined embeddings, and \emph{MIPS (true)}, which uses the true propensity scores of the pre-defined embeddings, which are not available in practice. 
For all experiments we report the MSE and standard errors averaged over 100 different subsampled datasets.

\begin{figure*}[tb]
    \centering
    \begin{subfigure}[t]{\linewidth}
        \centering
        \includegraphics[width=\textwidth]{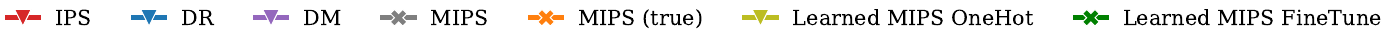}
    \end{subfigure}
    \begin{subfigure}[t]{0.48\linewidth}
        \centering
        \includegraphics[width=\textwidth]{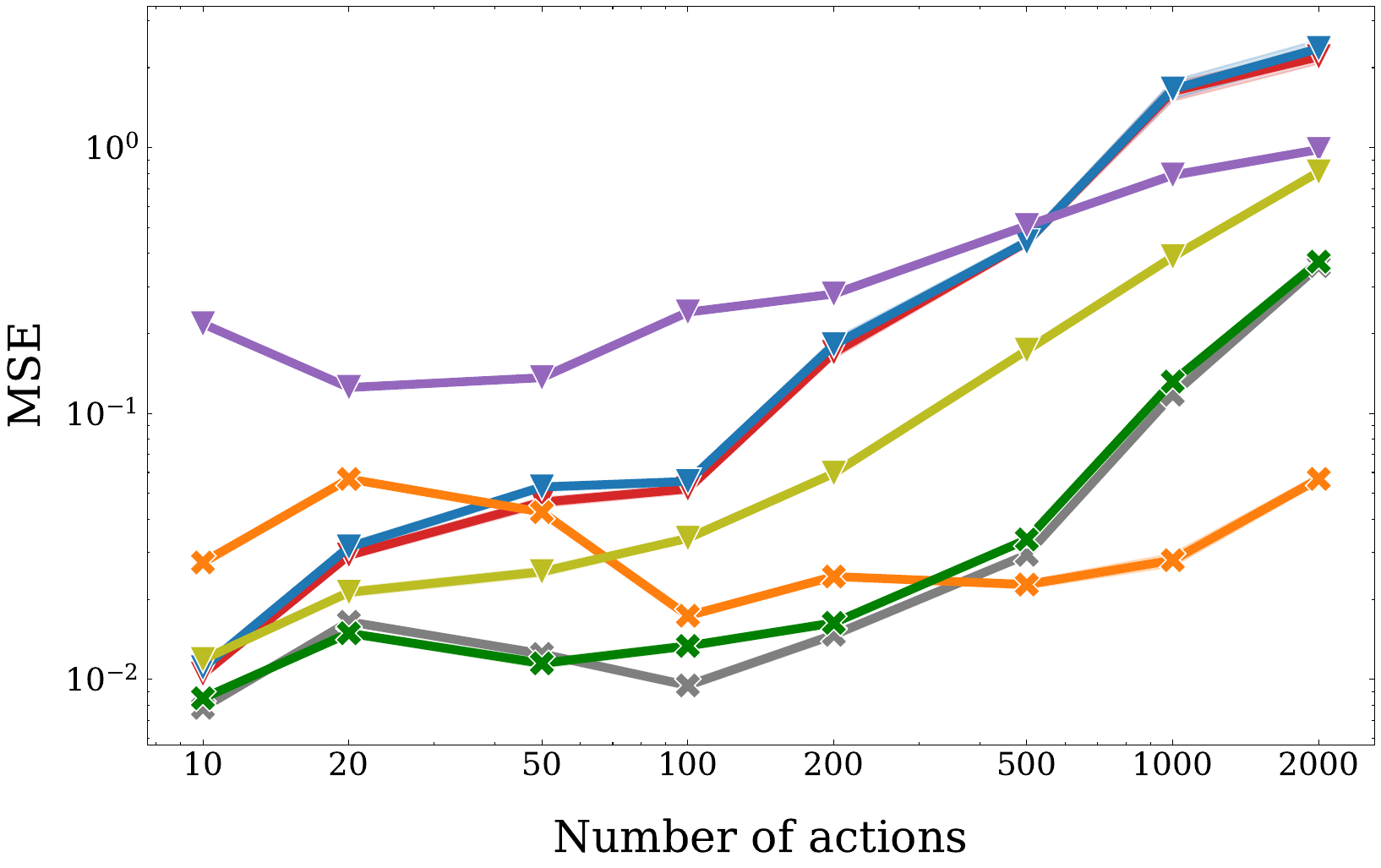}
    \end{subfigure}
    \hfill
    \begin{subfigure}[t]{0.48\linewidth}
        \centering
        \includegraphics[width=\textwidth]{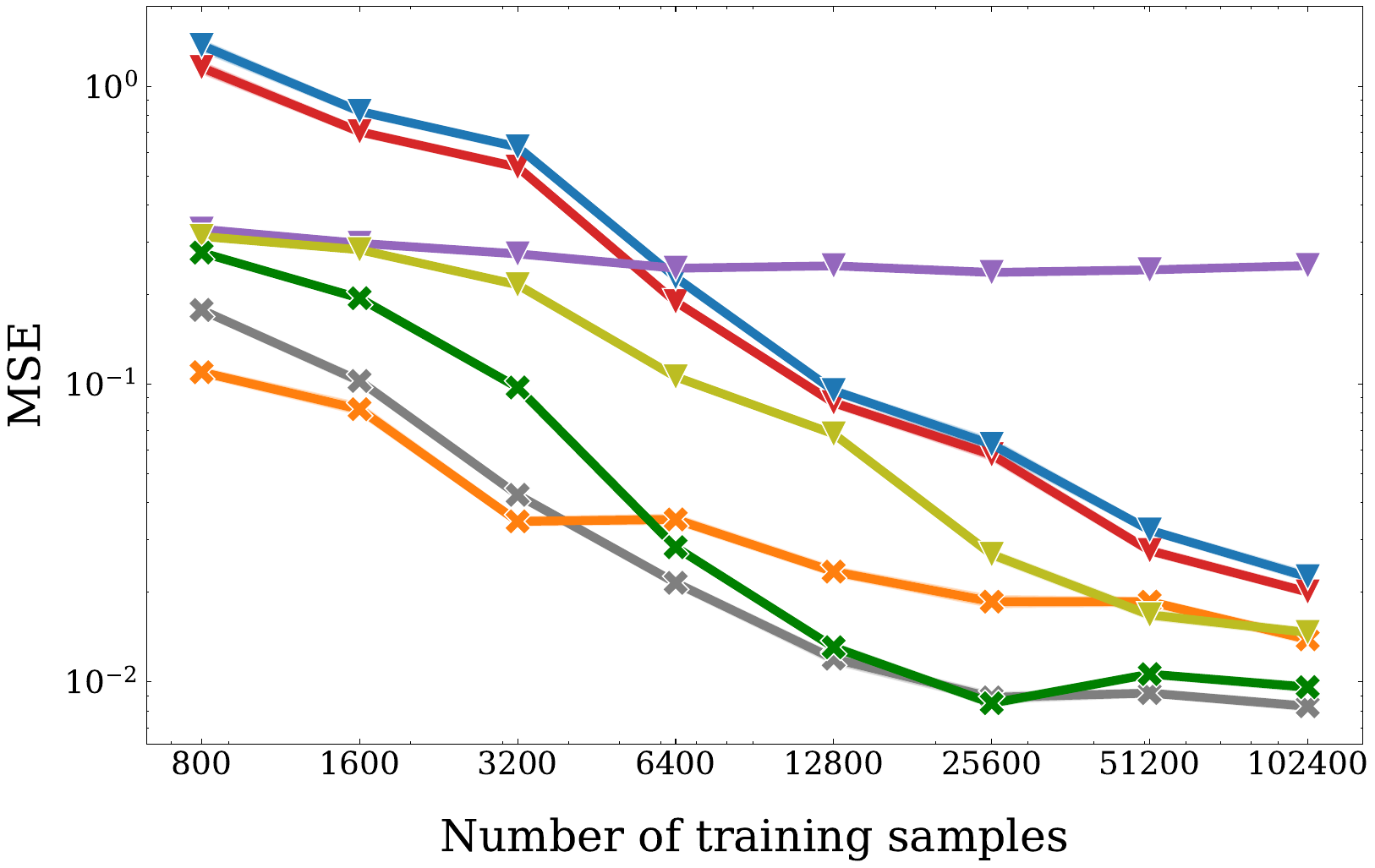}
    \end{subfigure}
    \caption{Synthetic experiments varying the number of actions and training samples. To better distinguish comparable methods, those marked with $\triangledown$ only use action identities, and those marked with $\times$ also use pre-defined action embeddings. \emph{Learned MIPS OneHot} outperforms all standard baselines. When we have enough data for every action, it performs just as well as IPS. As the variance grows with fewer samples per action, its error approaches the one of DM using the same model. The pre-defined embeddings ($d_\cE = 3$) have low bias and variance; hence our methods can not improve upon them. Shaded areas around the lines are standard errors (almost invisible).}
    \label{fig: synthetic linear}
    \vskip -0.15in
\end{figure*}

\paragraph{Can we improve upon standard baselines without using pre-defined embeddings?}
We evaluate our performance against all standard estimators in two sets of experiments with data generated according to the procedure described in \cref{sec: synthetic experiments}.
All pre-defined embedding dimensions are preserved, hence these embeddings are \emph{fully-informative}, i.e.\ $q(x, e) = q(x, e, a)$. 
We fix the number of pre-defined embedding dimensions to $d_\cE = 3$.
In the first experiment, we vary the number of actions from $10$ to $2{,}000$ with a fixed sample size of $n = 20{,}000$.
In the second experiment, we vary the number of samples in the training data $n = [800, 1600, \dots, 102400]$ and leave the number of actions fixed at $\abs{\cA} = 100$.
The results of these two experiments are presented in \cref{fig: synthetic linear}.

\emph{Learned MIPS OneHot} outperforms all other methods using action identities.
We can observe the difference between the two methods that use the same model, \emph{Learned MIPS OneHot} and \emph{DM}.
Using learned embedding propensities in 
MIPS results in a smaller bias compared to using them in direct reward prediction.
The error of \emph{Learned MIPS} keeps shrinking with more data. 
While IPS and DR do not pool information across actions, MIPS is using the embedding propensities influenced by every action, resulting in a better bias-variance trade-off.
When the number of actions is small enough, and we have enough data, IPS estimates already have a low variance, and we do not observe any significant improvements.

We also report the performance of the methods that use pre-defined embeddings.
Perhaps unsurprisingly, the learned embeddings perform worse than pre-defined embeddings, and even \emph{Learned MIPS FineTune} does not improve over them. 
This is because the pre-defined embeddings compress the information about the action well. 
In further experiments, we study more realistic benchmarks, where embeddings are higher-dimensional or do not fully capture the reward.
Both of these cases are common in practice, for example, when using pre-defined embeddings extracted from LLMs or when extending the setup to lists of items in learning to rank.

\begin{figure}[tb]
    \centering
    \begin{subfigure}[t]{\textwidth}
        \centering
        \includegraphics[width=\textwidth]{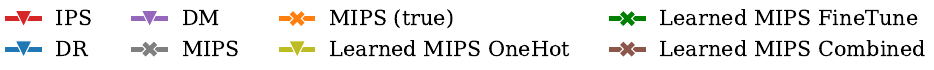}
    \end{subfigure}
    \begin{subfigure}[t]{0.7\textwidth}
    \includegraphics[width=\textwidth]{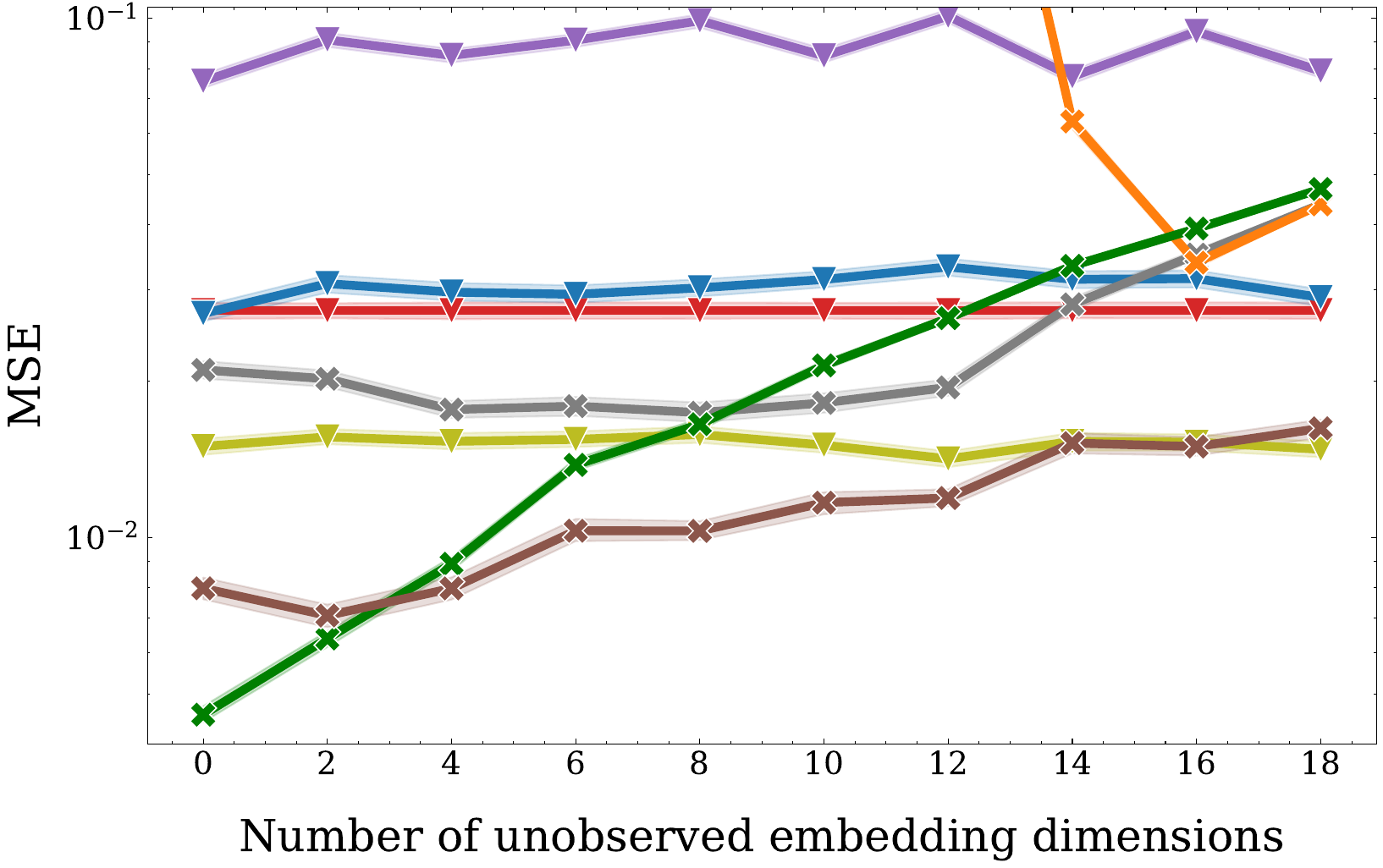}
    \end{subfigure}
    \caption{Synthetic experiments varying the number of unobserved dimensions. As we progressively hide more dimensions, pre-defined embeddings get more biased and methods using them get less accurate.  Combining \emph{Learned MIPS OneHot} and \emph{FineTune} yields the most robust results when the bias of pre-defined embeddings is unknown. Shaded areas around the lines are standard errors (almost invisible).}
    \label{fig: synthetic linear unobserved nuobsdim}
    \vskip -0.15in
\end{figure}

\paragraph{Can we improve upon the pre-defined embeddings?}
In the previous experiment, fine-tuned embeddings did not improve upon the pre-defined embeddings.
The reason is that the pre-defined embeddings have low bias and low variance.
This is rarely the case in practice.
In this experiment, we show that we can learn better embeddings when the pre-defined embeddings have a high number of embedding dimensions or when they do not capture the full effect on the reward.
We evaluate how the bias of the pre-defined embeddings influences our methods and whether it is better to learn the embeddings from action identities or fine-tune them from pre-defined embeddings in this setting.
We progressively add more bias similarly to \citet{saito_off-policy_2022}. 
We fix the number of dimensions in the action embedding to $d_\cE = 20$, and after the reward is generated, we hide a certain number of dimensions so the estimators do not have access to fully-informative embeddings.
The results are shown in \cref{fig: synthetic linear unobserved nuobsdim}.
As the embeddings worsen, our methods that learn or fine-tune them outperform MIPS.
We can see the bias-variance trade-off between \emph{Learned MIPS OneHot} and \emph{Learned MIPS FineTune}.
When the number of unobserved dimensions is small, the variance reduction gained from pre-defined embeddings is greater than induced bias. 
As we increase the number of unobserved dimensions, the bias of pre-defined embeddings can get even higher than the variance of IPS. 
In practice, we often do not know how biased the pre-defined embeddings are. 
Therefore, we introduce \emph{Learned MIPS Combined} and observe performance improvements when using this method for the majority of the values of the unobserved dimensions.
When the information in the embedding is complete (we do not hide any dimensions), providing additional dimensions as an action identity only makes the problem more difficult (\emph{Learned MIPS FineTune} outperforms \emph{Learned MIPS Combined} with more embedding dimensions). 
But as soon as some of the information in the embedding is omitted and the common \emph{embedding support} assumption \citep{saito_off-policy_2022} is violated, the action identity information is very useful, and \emph{Learned MIPS Combined} uses it to model the direct effect of the action missing from the pre-defined embedding.
As we show in the next section, it may be difficult to gauge in practice how much information is missing in the real-world embedding and which method would yield the best performance.
We leave the optimal method selection for future work.

\subsection{Real-world data}
\label{sec: real world experiments}

We follow the experimental setup of \citet{saito_off-policy_2022} to evaluate how the estimators perform on a real-world bandit dataset.
In particular, we use a fashion recommendation dataset that was collected on a large-scale fashion e-commerce platform, and comes in the form of $\cD = \set{(x_t, a_t, e_t, r_t)}_{t=1}^n$ tuples.
The dataset was collected during an A/B test under two policies: a uniform policy $\pi_0$ and a Thompson sampling policy $\pi$.
The size of the action space is $\abs{\cA} = 240$, and every action (a fashion item) comes with a pre-defined 4-dimensional action embedding $e$, which includes the item's price, brand, and hierarchical category information. 
We use $n = 10000$ sub-sampled observations from the ``ALL'' campaign in the dataset to be directly comparable to the prior work \citep{saito_off-policy_2022, peng_offline_2023}.

\begin{figure}[tb]
    \centering
    \begin{subfigure}[t]{\textwidth}
        \centering
        \includegraphics[width=\textwidth]{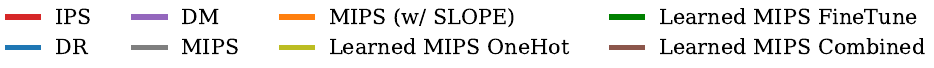}
    \end{subfigure}
    \begin{subfigure}[t]{0.7\textwidth}
        \includegraphics[width=\textwidth]{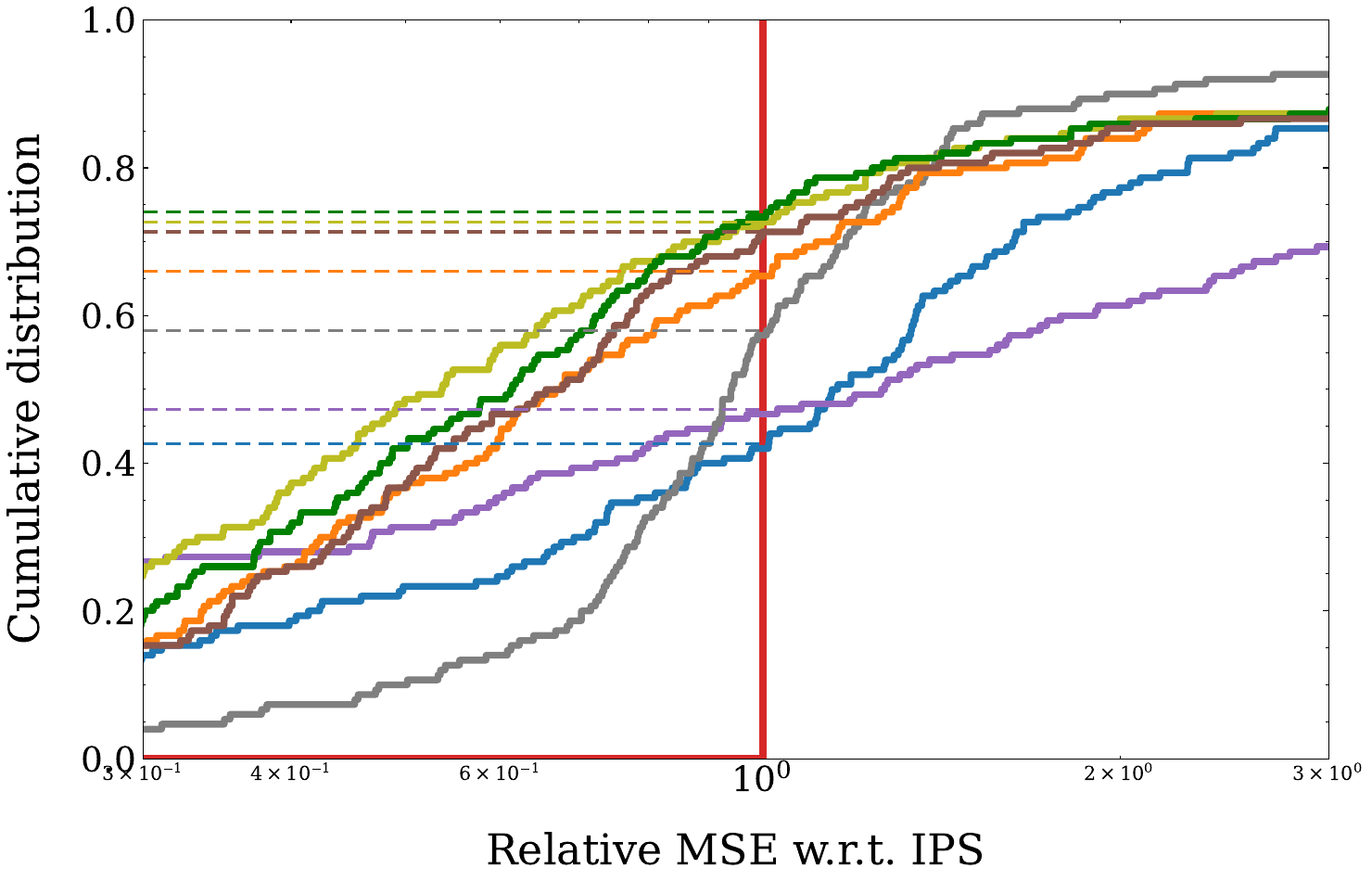}
    \end{subfigure}
    \caption{CDF of relative MSEs w.r.t\ IPS on the real-world dataset. The intersection of a method's curve with the IPS curve tells us the proportion of experiments in which the method performs better than IPS.}
    \label{fig: real world data}
    \vskip -0.15in
\end{figure}

In line with \citet{saito_off-policy_2022}, we repeat the experiment over 150 bootstrap samples, normalize the estimators' MSE relative to the MSE of IPS (i.e.\ divide the method's MSEs for a given data sample by the MSE of IPS for the same sample), and compute the cumulative distribution function (CDF) of these MSEs.
We report the results in \cref{fig: real world data}.
We compare against another competitive baseline \emph{MIPS(w/ SLOPE)} that greedily improves the performance by dropping embedding dimensions \citep{saito_off-policy_2022}. 
\emph{Learned MIPS} methods outperform IPS in about 75\% of runs, more often than any other baseline.
The next best-performing method is MIPS (w/ SLOPE), which outperforms IPS in about 65\% of runs.
For DM, we use the same model as in \emph{Learned MIPS OneHot}.
These results support our hypothesis that even though DM fails to model the reward, the embeddings it learns in the process can still be useful for model-free estimation.

\section{Conclusion}

In this work, we propose methods that assume a structured action space in the contextual bandit setting, and reduce the variance of model-free off-policy estimators by \emph{learning} action embeddings from the reward signal.
We extend MIPS to settings where pre-defined embeddings are not available, or are difficult to define.
At the same time, we show that the method can improve upon the pre-defined embeddings even when they are available.
The proposed method outperforms all standard baselines on synthetic and real-world fashion recommendation data, even if the reward model itself is inaccurate.
An interesting future direction is to study the embeddings learned by more complex classes of reward models, such as neural networks.
In the current experiments, we estimate the action distribution over the embeddings $p(a \mid e)$ using a discriminative model as proposed by \citet{saito_off-policy_2022}. 
To enable explicit bias-variance control in future work, it would be interesting to experiment with estimating these weights with a generative classifier and a \emph{prescribed} form of $p(a \mid e)$.

We would also like to extend our methods to the learning-to-rank setting, as list-level embeddings can dramatically reduce the combinatorial complexity of the problem.
This can be done by concatenating the item-level embeddings of a single list into a longer embedding and applying our method to it. 
We assume that in the same way our model exploits the similarity relationships between individual items, it would also learn to embed the browsing customer behavior impacting the final reward (e.g., the items at the end of the list are observed less frequently).
This would eliminate the need for explicitly modeling these behaviors by click models \citep{chuklin_click_2015}.
Another way to use our method in LTR, as it works with any model class, is to use some of the well-known list-wise LTR methods and use its intermediate output as the learned action embedding.

\subsubsection*{Acknowledgements}
We want to thank Mohamed Sadek for his contributions to the codebase.
The research conducted by Matej Cief (also with \href{https://slovak.ai/}{slovak.AI}) was partially supported by TAILOR, a project funded by EU Horizon 2020 under GA No. 952215, \url{https://doi.org/10.3030/952215}. 

\bibliographystyle{splncs04nat}
\bibliography{references}

\end{document}